\documentclass[letterpaper, 10 pt, conference]{ieeeconf}  %

\usepackage{graphicx}
\usepackage{subcaption} 

\usepackage{amssymb}
\usepackage{amsmath}
\usepackage{commath}
\usepackage{mathtools}

\usepackage{hyperref}
\usepackage{import}
\usepackage{paralist}
\usepackage{gensymb}
\usepackage{dsfont}
\usepackage{siunitx}
\usepackage[bottom]{footmisc}
\usepackage{framed}
\usepackage{balance} 
\usepackage{color, colortbl}
\usepackage[dvipsnames]{xcolor}
\usepackage[skins]{tcolorbox}
\usepackage{xurl}
\usepackage{tikzscale} %

\newsavebox{\measurebox}

\usepackage{amsmath,mathtools}

\newtcolorbox{myframe}[2][]{%
  enhanced,colback=white,colframe=black,coltitle=black,
  sharp corners,boxrule=0.4pt,left=0pt,right=0pt,top=0pt,bottom=0pt,
  fonttitle=\itshape,
  attach boxed title to top left={yshift=-0.3\baselineskip-0.4pt,xshift=2mm},
  boxed title style={tile,size=minimal,left=0.5mm,right=0.5mm,
  colback=white,before upper=\strut},
  title=#2,#1
}

\IEEEoverridecommandlockouts                              %

\DeclareMathOperator*{\minimize}{minimize}

\title{\LARGE \bf
Online Non-linear Centroidal MPC for  \\  Humanoid Robot Locomotion with Step Adjustment}

\author{Giulio Romualdi$^{1,2}$, Stefano Dafarra$^{1}$, Giuseppe L'Erario$^{1,3}$,\\
Ines Sorrentino$^{1,3}$, Silvio Traversaro$^{1}$ and Daniele Pucci$^{1,3}$%
\thanks{$^{1}$Artificial and Mechanical Intelligence, Istituto Italiano di Tecnologia, Genoa, Italy
        {\tt\footnotesize name.surname@iit.it}}%
\thanks{$^{2}$ DIBRIS, University of Genoa, Genoa, Italy}
\thanks{$^{3}$Machine Learning and Optimisation, University of Manchester, Manchester, UK}%
\thanks{The paper was supported by the Italian National Institute for Insurance against Accidents at Work (INAIL) ergoCub Project}
}

\begin{document}

\maketitle
\thispagestyle{empty}
\pagestyle{empty}

\begin{abstract}
This paper presents a Non-Linear Model Predictive Controller for humanoid robot locomotion with online step adjustment capabilities. The proposed controller considers the Centroidal Dynamics of the system to compute the desired contact forces and torques and contact locations. Differently from bipedal walking architectures based on simplified models, the presented approach considers the reduced centroidal model, thus allowing the robot to perform highly dynamic movements while keeping the control problem still treatable online. We show that the proposed controller can automatically adjust the contact location both in single and double support phases. The overall approach is then tested with a simulation of one-leg and two-leg systems performing jumping and running tasks, respectively. We finally validate the proposed controller on the position-controlled Humanoid Robot iCub. Results show that the proposed strategy prevents the robot from falling while walking and pushed with external forces up to 40 Newton for 1 second applied at the robot arm.
\end{abstract}

\section{Introduction}
Planning locomotion trajectories for humanoid robots requires to consider high-dimensional multi-body systems interacting with the surrounding environment. Furthermore, in the case of
unforeseen disturbances acting on the robot, it is necessary to adjust the location of the planned contacts to prevent the robot from falling. These adjustments can be generated, in a hierarchical fashion, by specialized controllers that provide proper joint references to the robot. This paper presents a control architecture that employs a reduced robot model in model predictive controllers to enable the step adjustment feature of humanoid robots. 

From a broader perspective, one can generate humanoid robot control actions by implementing instances of a popular architecture 
that exploits a model-based hierarchical architecture composed of three main layers \cite{doi:10.1142/S0219843616500079, Romualdi2019}. 
From outer to inner, the layers are named: \emph{trajectory optimization} layer, \emph{simplified model control} layer and the \emph{whole-body control} layer. The \emph{trajectory optimization} layer relies on simplified models such as the Linear Inverted Pendulum Model (LIPM)~\cite{Kajita} and the Divergent Component of Motion (DCM)~\cite{Englsberger2013} to generate feasible trajectories considering a nominal footstep sequence. The \emph{simplified model control} layer aims to generate a feasible CoM trajectory often combining the LIPM with Model Predictive Control (MPC) techniques~\cite{Camacho2007ModelControllers}, also known as the Receding Horizon Control (RHC)~\cite{Kajita2003}.
The \emph{whole-body control} generates desired joint values considering the complete robot model and the output of the other layers.
All  three layers use the robot state to compute their output. The inner the loop the higher the frequency at which the layer runs. When an external disturbance acts on the robot, it may be necessary to consider the robot state up to the trajectory optimization layer, recomputing the contact location to avoid the robot falling.
\begin{figure}[t]
  \centering
      \begin{subfigure}[b]{0.325\columnwidth}
        \centering
        \includegraphics[width=\textwidth]{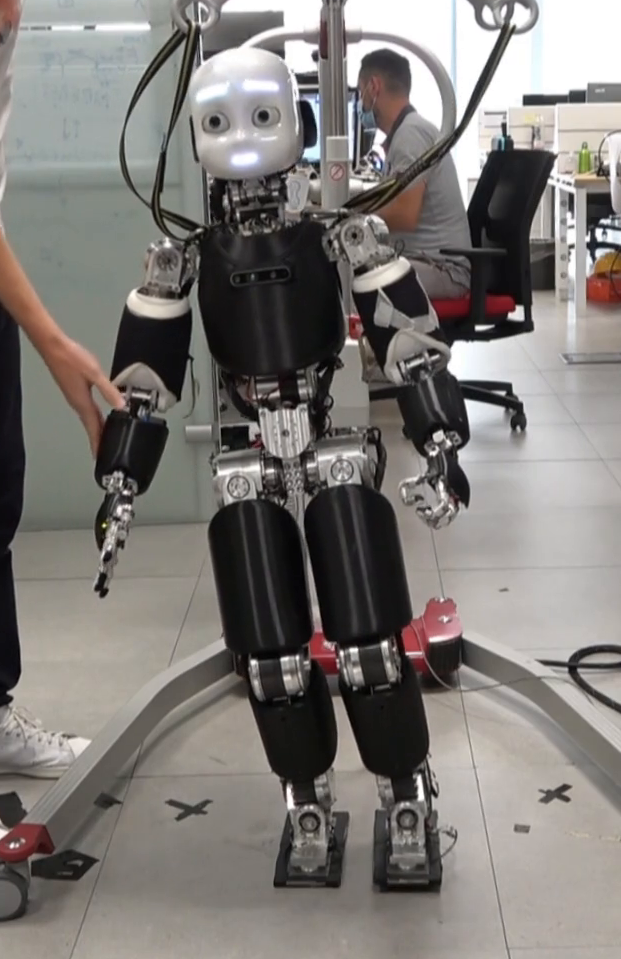}
    \end{subfigure}
    \hfill
     \begin{subfigure}[b]{0.325\columnwidth}
        \centering
        \includegraphics[width=\textwidth]{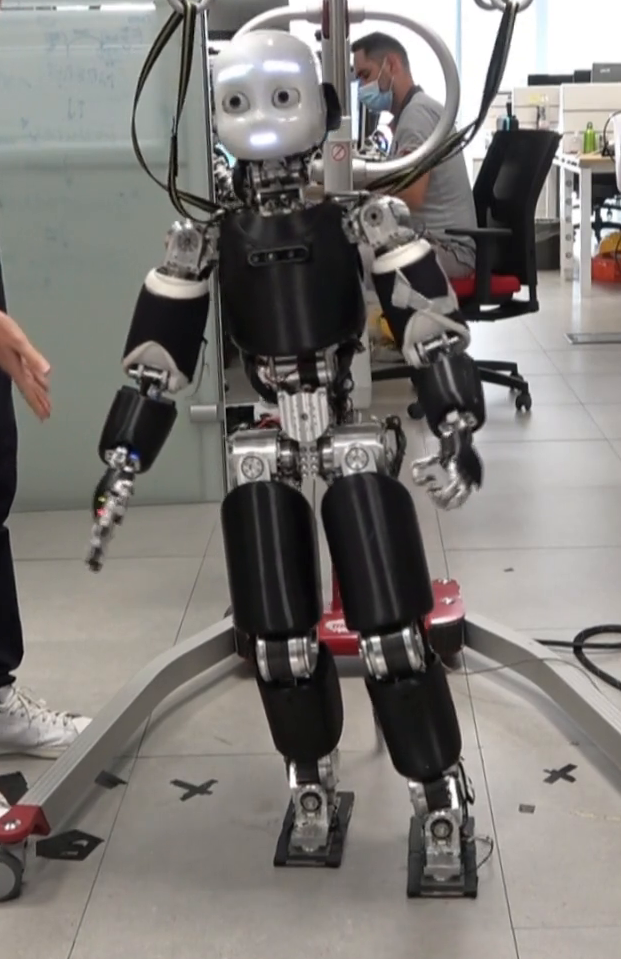}
    \end{subfigure}
    \hfill
    \begin{subfigure}[b]{0.325\columnwidth}
        \centering
        \includegraphics[width=\textwidth]{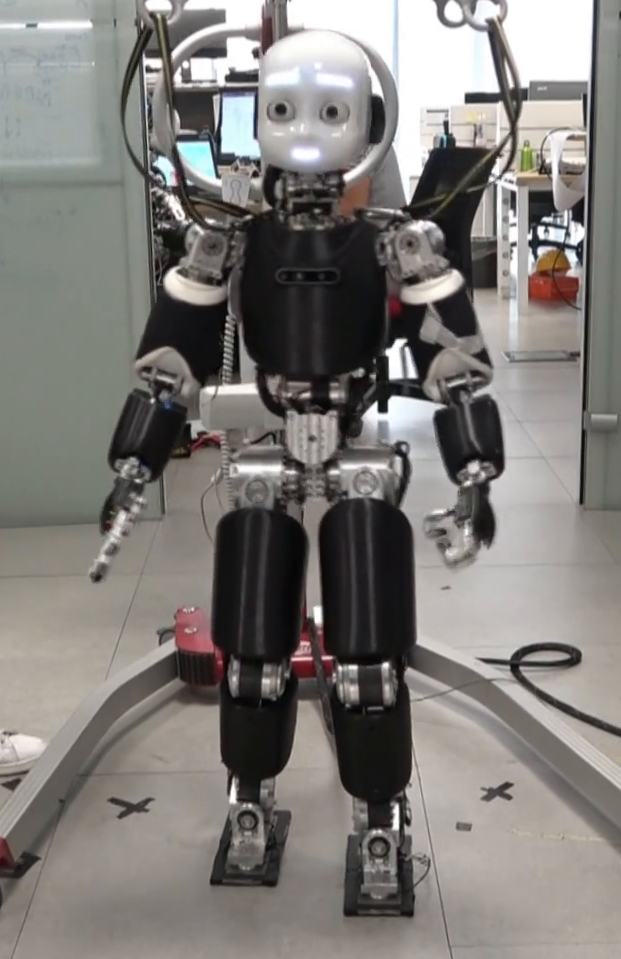}
    \end{subfigure}
  \caption{iCub reacts to the application of an external force. \label{fig:icub}}
\end{figure}
In recent years several attempts have been made in this direction.
The new contact locations can be optimized assuming constant step duration~\cite{Fengy2016RobustOptimization,Shafiee-Ashtiani2017,stephens2010pushforce} or adaptive step timing~\cite{Khadiv2016StepGaits, Shafiee2019OnlineRobots, Griffin2017WalkingAtlas,Scianca2020MPCFeasibility}.  
Approximating the robot with a simplified model allows solving the footsteps adjustment problem online. However, controller architectures based on  simplified models often treat the
contact adjustment strategy separately from the main control loop~\cite{Shafiee2019OnlineRobots,Mesesan2021OnlineLocomotion,Griffin2016DisturbanceRecovery} or apply heuristics-based strategies~\cite{DiCarlo2018DynamicControl}.
Furthermore, approaches based on simplified models require hand-crafted models for the task at hand~\cite{Kajita,Poulakakis2009TheHopper,Englsberger2015}, thus making the transition between different tasks, i.e. from locomotion to running, often very complex. 
\par
At the planning level, several attempts have been made for considering the robot Centroidal dynamics~\cite{Orin2013} and the robot kinematics to plan the desired motion trajectories~\cite{herzog2015trajectory,Fernbach2018CROC:Problem,Dafarra2020Whole-BodyApproach,Dai2015Whole-bodyKinematics}. 
With this approach, no prior knowledge is injected into the system to generate walking trajectories, but the whole-body motions result from a particular choice of the cost function.
While providing enhanced planning capabilities, given its complexity, a whole-body planner may require several minutes to compute a feasible solution. A common approach to reducing the computational demand is to assume a predefined contact sequence~\cite{Caron2017,carpentier2016versatile,Winkler2018GaitParameterization} while keeping the contact location and timings as an output of the planner. Even if such a choice simplifies the planning problem, the computational effort still prevents the use of the planner in a closed-loop controller.
Another common approach to reduce the computational complexity is to use an offline constructed gait library~\cite{Nguyen2020DynamicFunctions,Guo2021}. This approach simplifies the problem and allows the planner to run online. However, the offline gait can generate only a finite set of motions, thus rendering more complex the task of adding a new behavior to the library of motions. 
\par
\begin{figure*}[t]
    \centering
    \includegraphics{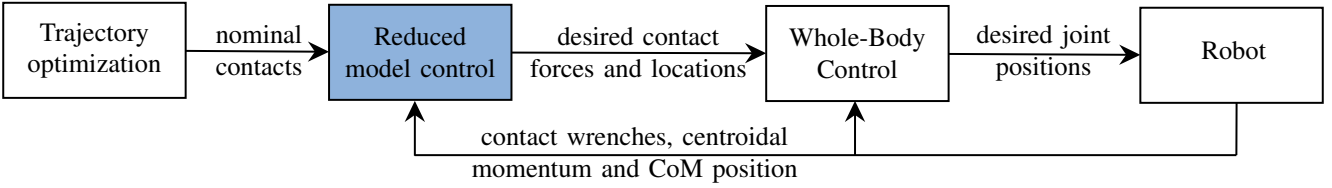}
    \caption{Reduced model control embedded in a three-layer walking architecture.}
    \label{fig:architecture}
\end{figure*}
This paper contributes towards the design of a non-linear Model Predictive Controller (MPC) that aims at generating online feasible contact locations and wrenches for humanoid robot locomotion. More precisely, differently from classical control architectures based on simplified models, the presented approach considers the reduced Centroidal dynamics model while keeping the problem still treatable online. By modeling the system using a reduced model instead of a simplified one, we achieve highly dynamic robot motions to be performed online. The contact location adjustment is considered in the Centroidal dynamics stabilization problem, thus it is not required to design an ad-hoc block for this feature. Furthermore, unlike existing work~\cite{Shafiee-Ashtiani2017, Jeong2019AStrategies}, our algorithm automatically considers double support phases, thus the contact adaptation feature is continuously active during both single and double support phases. We validate the proposed control strategy on a simulation of a one-leg, and two-leg systems, performing jumping and running tasks, respectively.  Results show that, differently from the simplified model controllers, the proposed Centroidal MPC generalizes on the number of contacts and the adjustment is automatically performed by the controller in the case of external disturbance acting on the system. Furthermore, we embed the MPC controller in a three-layer controller architecture -- see Fig.~\ref{fig:architecture} as a reduced-model control layer. The entire approach is also validated on the position-controlled Humanoid Robot iCub~\cite{Natale2017} -- see Fig.~\ref{fig:icub}. Results show that the proposed strategy prevents the robot from falling while walking and being pushed with external forces up to 40 Newton and lasting 1 second when applied at the robot arm.
\par
The paper is organized as follows. Sec.~\ref{sec:background} introduces the notation and recalls some concepts of floating base-systems used for locomotion. Sec.~\ref{sec:reduced_mpc} presents the Centroidal MPC. Sec~\ref{sec:results} presents the simulation results on different kinds of floating base systems and on the position-controlled Humanoid Robot iCub. Sec.~\ref{sec:conclusions} concludes the paper.

\section{Background}
\label{sec:background}
\subsection{Notation}
\begin{itemize}
\item $I_n$ and $0_n$ denote the $n \times n$ identity and zero matrices;
\item $\mathcal{I}$ denotes an inertial frame;
\item $\prescript{\mathcal{A}}{}{p}_\mathcal{C}$ is a vector that connects the origin of frame $\mathcal{A}$ and the origin of frame $\mathcal{C}$ expressed in the frame $\mathcal{A}$;
\item given $\prescript{\mathcal{A}}{}{p}_\mathcal{C}$ and $\prescript{\mathcal{B}}{}{p}_\mathcal{C}$,  $\prescript{\mathcal{A}}{}{p}_\mathcal{C} = \prescript{\mathcal{A}}{}{R}_\mathcal{B} \prescript{\mathcal{B}}{}{p}_\mathcal{C} + \prescript{\mathcal{A}}{}{p}_\mathcal{B}= \prescript{\mathcal{A}}{}{H}_\mathcal{B} \begin{bmatrix}
  \prescript{\mathcal{B}}{}{p}_\mathcal{C} ^\top & \;1
\end{bmatrix}^\top$. $\prescript{\mathcal{A}}{}{H}_\mathcal{B}$ is the homogeneous transformations and $\prescript{\mathcal{A}}{}{R}_\mathcal{B} \in SO(3)$ is the rotation matrix; 
\item the \emph{hat operator} is $^\wedge : \mathbb{R}^3  \to \mathfrak{so}(3)$, where $\mathfrak{so}(3)$ is the set of skew-symmetric matrices $x^\wedge y = x \times y$. $\times$ is the cross product operator in $\mathbb{R}^3$;
\item $\prescript{\mathcal{A}}{}{{v}}_\mathcal{B} \in \mathbb{R}^3$ is the time derivative of the relative position between the origin of the frame $\mathcal{B}$ and $\mathcal{A}$, $\prescript{\mathcal{A}}{}{{v}}_\mathcal{B}  = \prescript{\mathcal{A}}{}{\dot{p}}_\mathcal{B}$;
\item $\prescript{\mathcal{A}}{}{{f}}_\mathcal{B} \in \mathbb{R}^3$ denotes the force acting on the rigid body attached to the frame $\mathcal{B}$ expressed in $\mathcal{A}$;
\item $\prescript{\mathcal{A}}{}{\mu}_\mathcal{B} \in \mathbb{R}^3$ denotes the moment of a force about the origin of $\mathcal{B}$ expressed in $\mathcal{A}$;
\item the force acting on a point of a rigid body is uniquely identified by the wrench $\prescript{}{}{\mathrm{f}}_\mathcal{B}^ \top = \begin{bmatrix} \prescript{\mathcal{A}}{}{{f}}_\mathcal{B} ^ \top & \prescript{\mathcal{A}}{}{\mu}_\mathcal{B}^ \top \end{bmatrix}$;
\item $g$ is the gravity vector expressed in $\mathcal{I}$;
\item for the sake of clarity, the prescript $\mathcal{I}$ will be omitted.
\end{itemize}

\subsection{Floating base multi-body system}
A floating base multi-body system is composed of $n+1$ links connected by $n$ joints with one degree of freedom each. Since none of the system links has a fixed pose with respect to (w.r.t.) the inertial frame $\mathcal{I}$, its configuration is defined by considering both the joint positions $s$ and the homogeneous transformation from the inertial frame to the robot base frame, $\mathcal{B}$. The configuration of the robot is identified by the triplet $q = (\prescript{\mathcal{I}}{}{p}_\mathcal{B}, \prescript{\mathcal{I}}{}{R}_\mathcal{B}, s) \in  \mathbb{R}^3 \times SO(3) \times \mathbb{R}^n$.
The velocity of the system is determined by the triplet $ \nu = (\prescript{\mathcal{I}}{}{v}_\mathcal{B}, \prescript{\mathcal{I}}{}{\omega}_\mathcal{B}, \dot{s})$, where $\dot{s}$ is the time derivative of the  joint positions.
\par

The centroidal momentum ${}_{\bar{G}} h ^\top= \begin{bmatrix}
    {}_{\bar{G}} h^{p\top} & {}_{\bar{G}} h^{\omega\top}
\end{bmatrix} \in \mathbb{R}^6$  is the aggregate linear and angular momentum of each link of the robot referred to the robot center of mass (CoM). The vectors ${}_{\bar{G}} h^p \in \mathbb{R}^3$ and ${}_{\bar{G}} h^\omega \in \mathbb{R}^3$ are the linear and angular momentum, respectively. The coordinates of ${}_{\bar{G}} h$ are expressed w.r.t. a frame centered in the robot CoM and oriented as the inertial frame $\mathcal{I}$~\cite{Orin2013}. The time derivative of the centroidal momentum depends on the external contact wrenches acting on the system~\cite{nava16}, thus leading to:
\begin{equation}
	    {}_{\bar{G}} \dot{h} = \sum_{i = 1}^{n_c}\begin{bmatrix}
      I_3 & 0_3 \\
      (p_{\mathcal{C}_i} - p_{\text{CoM}})^\wedge & I_3 
    \end{bmatrix} \mathrm{f}_i + m\bar{g},
    \label{eq:centroidal_dynamics_original}
\end{equation}
where $\bar{g}^\top = \begin{bmatrix} g^\top & 0_{1\times3} \end{bmatrix}$, $m$ is the mass of the robot and $n_c$ the number of active contacts.
The relation between the linear momentum ${}_{\bar{G}} h^{p}$ and the robot CoM velocity is linear and depends on the robot mass $m$, i.e. ${}_{\bar{G}} h^{p} = m v_{\text{CoM}}$.

Consider a rigid body in contact with a rigid environment and assume that the contact surface is described by a rectangle. Then, the contact wrench acting on the rigid body is uniquely described by four pure forces acting on the corner of the contact surface~\cite{Caron2015,Caron}. Indeed in the case of a rectangular contact surface, twelve variables define the six-dimensional wrench. Thanks to this choice, several contact configurations can be modeled independently, depending on the number of points in contact~\cite{Dafarra2020Whole-BodyApproach,Dai2015Whole-bodyKinematics}. Given the relation between pure forces and contact wrench, the centroidal dynamics \eqref{eq:centroidal_dynamics_original} can be rewritten as:
\begin{equation}
 	{}_{\bar{G}} \dot{h} = \sum_{i = 1}^{n_c} \sum_{j = 1}^{n_v} \begin{bmatrix}
       I_3 \\
       (p_{\mathcal{C}_i} + R _{\mathcal{C} _ {i} }p_{v_{i,j}} - p_{\text{CoM}})^\wedge
     \end{bmatrix} f_{i,j} + m\bar{g}.
     \label{eq:centroidal_dynamics}
\end{equation}
The number of vertices of to the contact surface is denoted by $n_v$, while $p_{v_{i,j}}$ is the position of the vertex $j$ of the contact $i$ expressed w.r.t the frame associated with the contact surface. $f_{i,j}$ is the pure force applied to the vertex $j$ of the contact $i$.
\par
Assume a rigid body interacting with the environment. The contact force is supposed to be non-null only if the point is in contact with the environment. The condition is called \emph{complementary condition} and writes as:
\begin{equation}
    h(p_{C_i})n(p_{C_i})^\top f_i = 0,
    \label{eq:complementary_condition}
\end{equation}
where $h$ computes the distance between the point $p_{C_i}$ and the environment and $n(p_{C_i})$ returns the direction normal to the contact surface at the point $p_{C_i}$.

\section{Centroidal Model Predictive Controller}
\label{sec:reduced_mpc}

Let us assume the presence of a high-level contact planner that generates only the contact location and timings, e.g.~\cite{8594277}.
The reduced model control objective is to implement a control law that generates feasible contact wrenches and locations while considering the Centroidal dynamics of the floating base system, and a nominal set of contact positions and timings. The control problem is formulated using the Model Predictive Control (MPC) framework. 

The control objective is achieved by framing the controller as a constrained optimization problem composed of three main elements, namely: the \emph{prediction model}, an \emph{objective function} and a set of \emph{inequality constraints}.

The complete MPC formulation is presented in Sec.~\ref{sec:mpc_formulation}. 

\subsection{Prediction Model}
\label{sec:dynamics}

Since the proposed controller assumes the knowledge of the contact sequence, it is possible to define the variable $\Gamma_i \in \{0, 1\}$ for each contact. $\Gamma_i$ represents the contact state at a given instant. $\Gamma_i(t) = 0$ indicates that the contact $i$-th is not active at the time $t$, while, when $\Gamma_i(t) = 1$ the contact is active. 
Thanks to this assumption, it is not necessary to introduce the contact force complementary condition~\eqref{eq:complementary_condition}. In fact, considering the complementary condition in an optimization algorithm may cause problems for the nonlinear optimization solvers because the constraint Jacobian gets singular, thus violating the linear independence constraint qualification (LICQ) on which most solvers rely upon~\cite{BettsPractical2010}.
As a consequence of the introduction of $\Gamma_i$, \eqref{eq:centroidal_dynamics} rewrite as
\begin{IEEEeqnarray}{cl}
\IEEEnonumber
    	{}_{\bar{G}} \dot{h} &=\sum_{i = 1}^{n_c} \sum_{j = 1}^{n_v} \begin{bmatrix}
      I_3 \\
      (p_{\mathcal{C}_{i}} + R _{\mathcal{C} _ {i}}p_{v_{i,j}} - p_{\text{CoM}})^\wedge
    \end{bmatrix} \Gamma_i f_{i,j} + m\bar{g}\\
     \IEEEyesnumber
    &=\mathcal{F}\left(p_{\mathcal{C}_i}, p_{\text{CoM}}, f_{i,j}\right).
    \label{eq:centroidal_dynamics_discretized}
\end{IEEEeqnarray}
In $\mathcal{F}$, we explicitly express the dependency on the unknown variables $p_{\mathcal{C}_i}$,  $p_{\text{CoM}}$ and $f_{i,j}$.
\par
Since the MPC aims to compute the control outputs online, the optimal control problem formulation should be as general as possible on the number of the active contact phases in the prediction windows. For this reason, we consider each contact location as a continuous variable subject to the following dynamics
\begin{equation}
    \dot{p} _{\mathcal{C} _ i}= ( 1 - \Gamma _ i)  v_{\mathcal{C} _ i},
    \label{eq:contact_dynamics}
\end{equation}
where $v _ {\mathcal{C}_i}$ is the contact velocity. To give the reader a better comprehension of \eqref{eq:contact_dynamics}, we can imagine that when the contact is active, i.e. $\Gamma_i = 1$, Eq.~\eqref{eq:contact_dynamics} becomes $\dot{p} _{\mathcal{C} _ i} = 0$. In other words,  the contact location is constant if the contact is active.

\subsection{Objective function \label{sec:tasks_formulation}}
The objective function is defined in terms of several tasks. A detailed explanation of each task follows.

\subsubsection{Contact force regularization}

Each contact link is subject to the effect of different contact forces. Since the net effect is given by the sum of all these forces, we want them to be as similar as possible. As a consequence, we add a task that weighs the difference of each contact force from the average, for a given contact link:
\begin{equation}
    \Psi_{f_{i,j}} = \frac{1}{2} \left\| \frac{1}{n_v}\sum_{w = 1}^{n_v} f_{i,w} - f_{i,j} \right\|^2_{\Lambda_{f_{i,j}}},
\end{equation}
where $\Lambda_{f_{i,j}}$ is a positive definite diagonal matrix. 
\par
To reduce the rate of change of the contact force, we minimize the contact force derivative by considering the following task:
\begin{equation}
    \Psi_{\dot{f}_{i,j}} = \frac{1}{2} \left\|  \dot{f}_{i,j} \right\|^2_{\Lambda_{\dot{f}_{i,j}}},
\end{equation}
where $\Lambda_{\dot{f}_{i,j}}$ is a positive defined diagonal matrix.

\subsubsection{Centroidal dynamics task}
To follow a desired centroidal momentum trajectory, we minimize the weighted norm of the error between the robot centroidal quantities and the desired trajectory:
\begin{equation}
    \Psi_{h} = \frac{1}{2} \left\| \prescript{}{\bar{G}}{h}^{\omega ^n} - \prescript{}{\bar{G}}{h}^\omega \right\|^2_{\Lambda_{h}} + \frac{1}{2} \left\| p_{\text{CoM}} ^ n -p_{\text{CoM}}  \right\|^2_{\Lambda_{\text{CoM}}}, 
    \label{eq:task_centroidal}
\end{equation}
where $\Lambda_h$ and $\Lambda_{\text{CoM}}$ are two positive definite diagonal matrices. The desired angular momentum $\prescript{}{\bar{G}}{h}^{\omega ^n}$  and CoM position $ p_{\text{CoM}} ^ n$, should be seen as regularization terms. In this work $\prescript{}{\bar{G}}{h}^{\omega ^n}$ is always set equal to zero, while $ p_{\text{CoM}} ^ n$ is a 5th-order spline passing through the nominal contact locations, whose initial and final velocity and acceleration are zero.

\subsubsection{Contact Location regularization}
To reduce the difference between the nominal contact location and the one computed by the controller, the following task is considered:
\begin{equation}
    \Psi_{p_{\mathcal{C}_i}} = \frac{1}{2} \left\| p_{\mathcal{C}_i^n}  - p_{\mathcal{C}_i}  \right\|^2_{\Lambda_{p_{\mathcal{C}_i}}}.
    \label{eq:task_contact}
\end{equation}
Here $p_{\mathcal{C}_i^n}$ is the nominal contact position provided by a high-level planner and $\Lambda_{p_{\mathcal{C}_i}}$ is a positive definite diagonal matrix.

\subsection{Inequality Constraints}
\label{sec:constraints}
\subsubsection{Contact Force Feasibility}
To be feasible, the contact force should belong to a second-order cone, i.e. a Lorentz cone \cite{Lobo1998ApplicationsProgramming}. However, for the sake of simplicity, the friction cone is often approximated by the conic combination of $n$ vectors. The half-space representation of the friction cone approximation is given by a set of linear inequalities of the form $A R ^\top_{\mathcal{C} _ {i} } f _ {i, j} \le b$. Where $A$ and $b$ are constants and depend on the static friction coefficient. 
\subsubsection{Contact location constraint}
\begin{figure}[t]
    \includegraphics{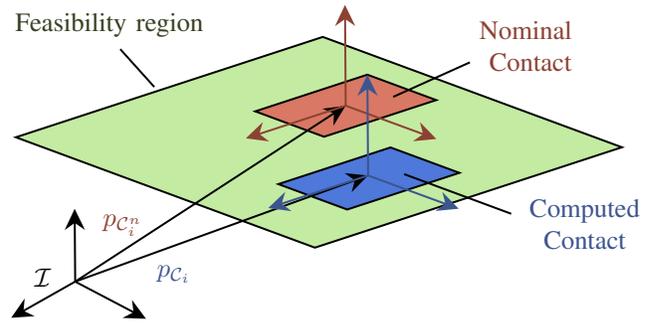}
    \caption{The contact feasibility region.}
    \label{fig:contact_adjustment}
\end{figure}

The proposed controller aims to compute the contact location, in particular, the contact position should belong to the feasibility region described by a rectangle containing the nominal contact location - Fig.~\ref{fig:contact_adjustment}.
The contact location constraint is described by:
\begin{equation}
    l_b\le R_{\mathcal{C} _ {i} } ^\top  (p_{\mathcal{C}_i ^n} - p_{\mathcal{C}_i}) \le u_b,
\end{equation}
where $l_b$ and $u_b$ are the lower and upper bound of the rectangle represented in the frame attached to the contact.

\subsection{MPC formulation}
\label{sec:mpc_formulation}
Combining the set of tasks (Sec.~\ref{sec:tasks_formulation}), with the prediction model (Sec. ~\ref{sec:dynamics}) and the inequality constraints  (Sec.~\ref{sec:constraints}), we formulate the MPC as an optimization problem. 
\par
The MPC problem is solved using a Direct Multiple Shooting method with a constant sampling time $T$ \cite{BettsPractical2010}. The controller outputs are generated using the Receding Horizon Principle \cite{Mayne90MPC}, adopting a fixed prediction window with a
length equal to $N$ samples.

The MPC formulation is finally obtained solving the following optimization problem
\begin{IEEEeqnarray*}{CL}
	\nonumber
	\minimize_{\substack{\mathcal{X}_k, \mathcal{U}_k, \\ k = [0, N]}}& \sum_{k = 0} ^ N \left(\sum_{i,j} \Psi_{f_{i,j}} + \sum_{i, j} \Psi_{\dot{f}_{i,j}} + \Psi_h + \sum_i \Psi_{p_{\mathcal{C}_i}}\right)
 \label{costFunction}\\
	\text{subject to } & {}_{\bar{G}} h[k + 1] = \mathcal{F}\left(p_{\mathcal{C}_i}, p_{\text{CoM}}, f_{i,j}\right) T + {}_{\bar{G}} h[k] \\
	& p_{\text{CoM}}[k+1] = \frac{{}_{\bar{G}} h^p[k]}{m} T  + p_{\text{CoM}}[k] \\
	& p_{\mathcal{C} _ i}[k + 1] = p_{\mathcal{C} _ i}[k] + T ( 1 - \Gamma _ i[k]) v_{\mathcal{C} _ i}[k] \\
	& A R ^\top_{\mathcal{C} _ {i} } [k] f _ {i, j}[k] \le b \\ 
	&     l_b\le R_{\mathcal{C} _ {i} } ^\top[k]  (p_{\mathcal{C}_i^n} [k] - p_{\mathcal{C}_i}[k]) \le u_b .
\end{IEEEeqnarray*}
Here $\mathcal{X}_k$ and $\mathcal{U}_k$ contain respectively the controller state and output at a time instant $k$:
\begin{equation}
    \mathcal{X}_k ^\top = \begin{bmatrix} p_{\text{CoM}}[k]^\top & \prescript{}{\bar{G}}{h}[k]^\top &  p_{\mathcal{C}_i}[k]^\top \end{bmatrix},
\end{equation}
\begin{equation}
    \mathcal{U}_k ^\top = \begin{bmatrix} f_{i,j}[k]^\top &  v_{\mathcal{C}_i}[k]^\top \end{bmatrix}.
\end{equation}

\begin{figure*}
    \centering
    \begin{myframe}{One-leg jumping robot}
    \includegraphics[width=\textwidth]{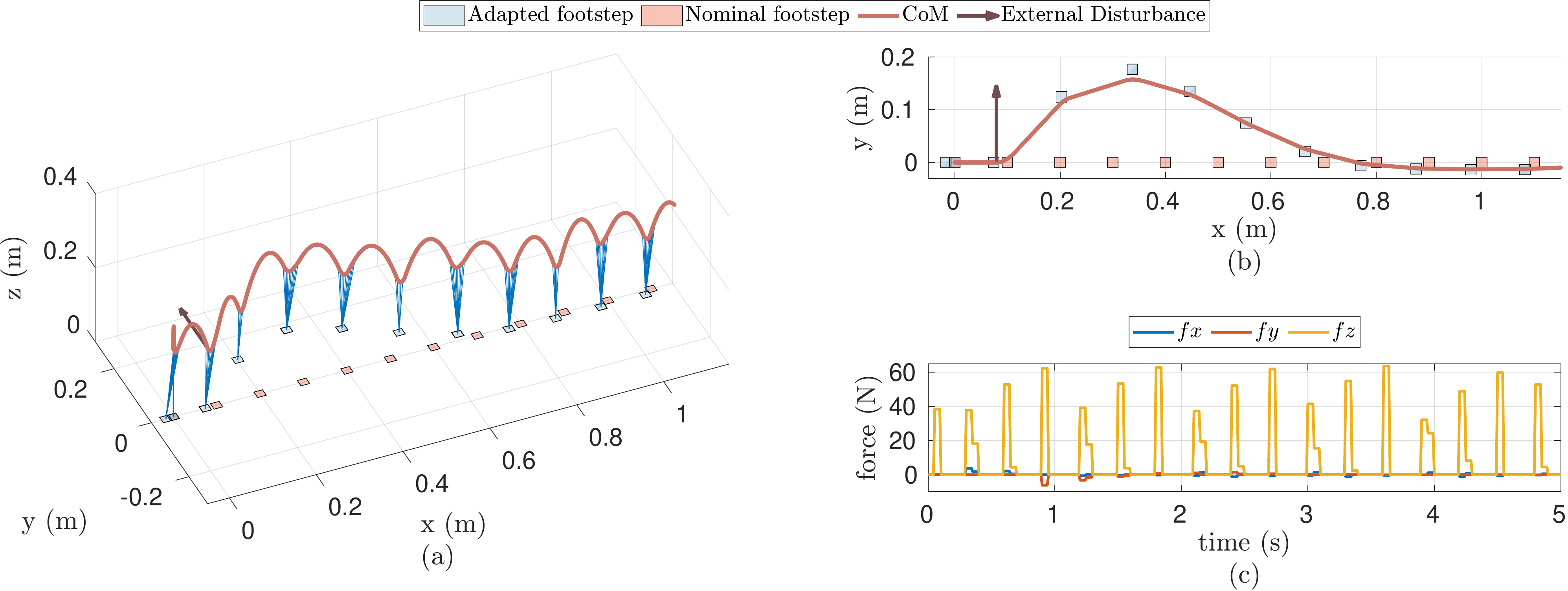}
    \end{myframe}
    \caption{(a)-(b) Trajectories generated by the MPC on a one-leg robot performing a jumping task. (c) Desired contact forces.}
    \label{fig:one_leg_jumping}
\end{figure*}
Since the centroidal dynamics \eqref{eq:centroidal_dynamics} is a non-linear non-convex function, the optimizer may find a local minimum. This may
result in a sub-optimal solution for the given task. As a consequence, the initialization of the solver may play a crucial role to drive the optimizer to the desired solution.
In our case, the CoM is initialized with the nominal CoM trajectory $p_{\text{CoM}}^n$, while all the other variables are set to zero.

\begin{figure*}
    \centering
    \begin{myframe}{Two-legs running robot}

    \includegraphics[width=\textwidth]{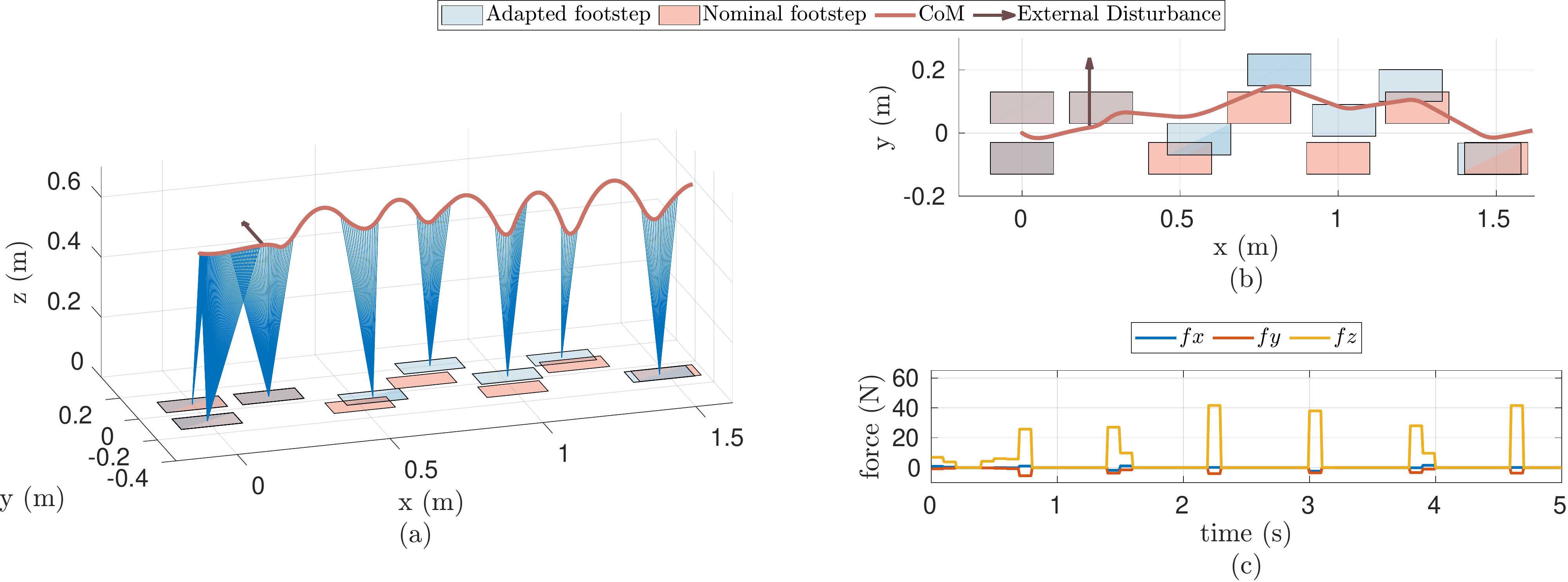}
    \end{myframe}
    \caption{(a)-(b) Trajectories generated by the MPC on a two-legs robot performing a running task. (c) Desired contact forces.}
    \label{fig:running_traj}
\end{figure*}

\begin{figure*}
    \centering
    \begin{myframe}{iCub walking}
    \includegraphics[width=0.945\textwidth]{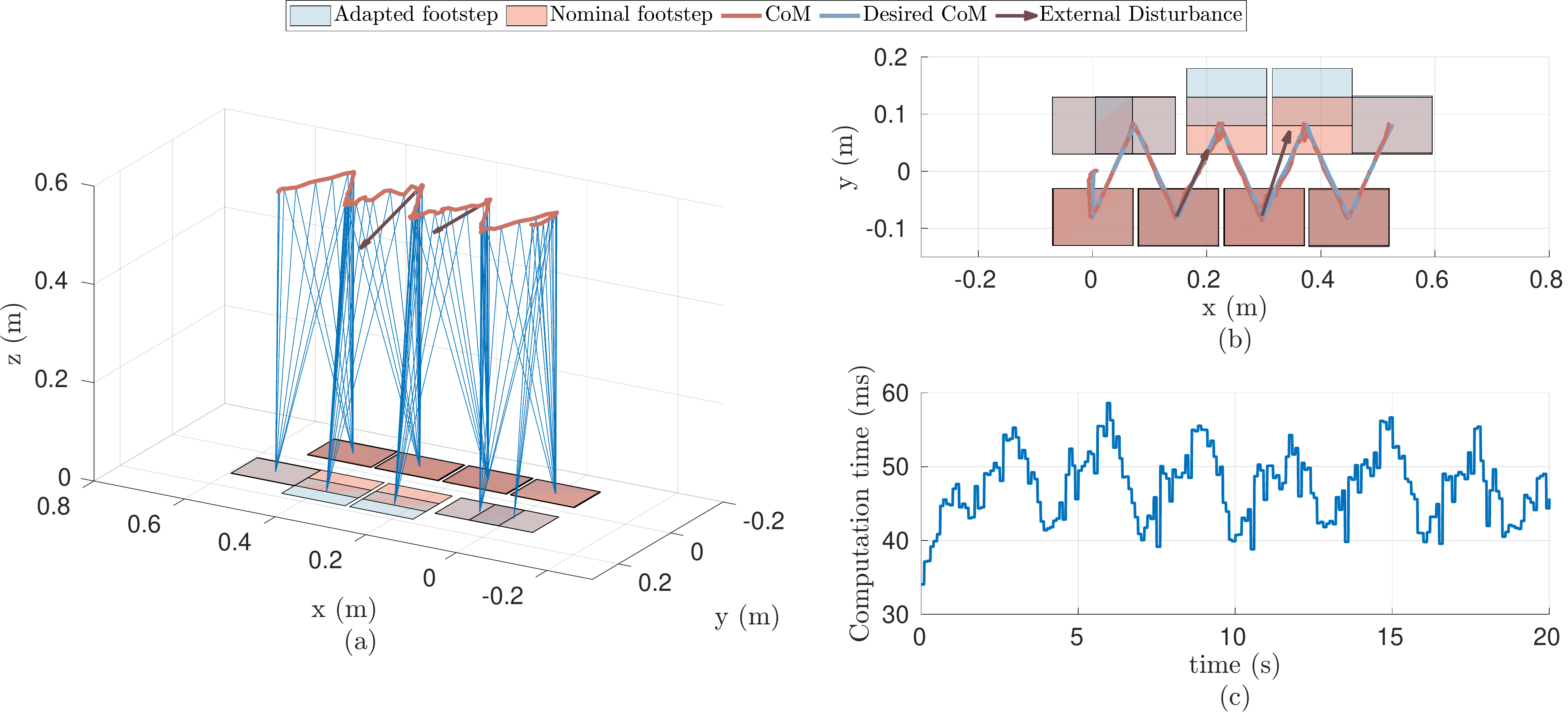}
    \end{myframe}
    \caption{(a)-(b) Trajectories generated by the three-layer controller architecture on the iCub robot. (c) Computation time.}
    \label{fig:icub_walking}
\end{figure*}

\section{Results}
\label{sec:results}
In this section, we present the validation results of the control strategy presented in Sec.~\ref{sec:reduced_mpc}.
CasADi \cite{Andersson2018CasADi:Control} and IPOPT 3.13.4~\cite{Wachter2005OnProgramming} with HSL\_MA97~\cite{Hogg2011HSL_MA97Systems} libraries are used to solve the non-linear optimization problem. The code is available at~\url{https://github.com/ami-iit/paper_romualdi_2022_icra_centroidal-mpc-walking}.
\par
To validate the performances of the proposed control we present two main experiments. First, we test the Centroidal MPC considering only the Centroidal Dynamics~\eqref{eq:centroidal_dynamics} for a one-leg and two-legs floating base systems. Secondly, we present the results obtained with the implementation of the control architecture shown in Fig.~\ref{fig:architecture} on an improved version of the humanoid robot iCub~\cite{Natale2017}. 
In both the scenarios we analyze the performances of the controller while running on a $10^\text{th}$ generation Intel$^\text{\textregistered}$ Core i7-10750H laptop equipped with Ubuntu Linux 20.04. 
\subsection{Reduced Models simulation}
Fig.~\ref{fig:one_leg_jumping} depicts the trajectory generated by the Centroidal MPC in the case of a floating base system equipped with one leg. The system has a mass of $\SI{1}{\kilo \gram}$ and the foot is approximated by a point, i.e. $n_c = 1$ and $n_v=1$ in Eq.~\eqref{eq:centroidal_dynamics}. The MPC takes (in average) less than \SI{20}{\milli \second} for evaluating its output. At $t \approx \SI{1}{\second}$ an external force of magnitude $\SI{5}{\newton}$ acts for $\SI{0.5}{\second}$ at the system CoM.
The MPC automatically compensates the disturbance effect by adapting the location of the footstep with an average of $\SI{10}{\centi \meter}$ - Figs.~\ref{fig:one_leg_jumping}a and \ref{fig:one_leg_jumping}b.
Fig.~\ref{fig:one_leg_jumping}c shows the contact force computed by the controller.
\par
Fig.~\ref{fig:running_traj} presents the trajectory generated by the Centroidal MPC in the case of a floating base system equipped with two legs. The system weighs $\SI{1}{\kilo \gram}$ and it has a foot length and width of  $\SI{20}{\centi \meter}$ and $\SI{10}{\centi \meter}$, respectively. In this case, $n_c = 2$ and $n_v=4$ in Eq.~\eqref{eq:centroidal_dynamics}. The MPC takes (in average) less than \SI{80}{\milli \second} for evaluating its output. 
In this experiment, we analyze the capabilities of the MPC in the transition from locomotion to running. 
At $t \approx \SI{1}{\second}$ the planned contact sequence switches from a bipedal locomotion pattern, where a single support phase is always preceded by a double support one, to a running pattern, where the single support phases are followed by aerial phases. Furthermore at $t \approx \SI{1.5}{\second}$ an external force of magnitude $\SI{5}{\newton}$ acts for $\SI{0.5}{\second}$ at the robot CoM.
The MPC handles the transition from locomotion to running while dealing with the external disturbance effect. Figs.~\ref{fig:running_traj}a and \ref{fig:running_traj}b show, in blue, the optimal contact location computed by the controller. The distance between the nominal contact and the computed one is (in average) $\SI{6}{\centi\meter}$.
Finally, Fig.~\ref{fig:running_traj}c shows the contact force computed by the controller.

\subsection{Test on the iCub Humanoid Robot}
To validate the performances of the Centroidal MPC on humanoid robots, we attached the controller to the three-layer architecture presented in~\cite{Romualdi2019} -- Fig.~\ref{fig:architecture}.
In this scenario, the \emph{trajectory optimization layer} is in charge of generating the nominal contact locations and timings. The nominal contact pose is considered as a regularization term for the contact position~\eqref{eq:task_contact} and to compute the regularized CoM trajectory~\eqref{eq:task_centroidal}.
The \emph{Centroidal MPC} generates the feasible contact wrenches for the current active contacts and the new location of the future active contacts. The future contact location is then set in the swing foot trajectory planner to generate a smooth trajectory for the foot. 
Finally, the inner \emph{whole-body control} loop evaluates the robot generalized velocity $\nu$, which is the solution to a stack of tasks formulation with hard and soft constraints considering the references computed by the Reduced model control layer.
Here, the tracking of the feet and the CoM trajectories are considered as high priority tasks while the torso orientation is treated as a low priority task. Furthermore, to attempt the stabilization of the zero-dynamics of the system, a postural term is added as a low-priority task.
Since the decision variable is the robot generalized velocity $\nu$ and tasks depend, through the Jacobian matrices, linearly on $\nu$ the optimization problem can be treated as a Quadratic Programming (QP) problem and then solved via an off-the-shelf solver.
The joint velocities $\dot{s}$ included in the solution of the above problem, are then integrated to get joint position references for the low-level position controller. 
In our implementation, the whole-body control layer takes (in average) less than $\SI{1}{\milli \second}$ for evaluating its outputs. The OSQP library~\cite{Stellato2018b} solves the QP problem.
\par
To analyze the step recovery capabilities of the whole architecture, the robot is perturbed with an external force acting on the right arm while walking. Since the robot is position control, it behaves rigidly when the external force is applied. Consequentially, the position of the CoM is not perturbed. To mitigate this effect, the estimated external force is considered as a measured disturbance in the MPC.
\par
The MPC takes less than \SI{60}{\milli \second} for evaluating its output -- Fig.~\ref{fig:icub_walking}c. 
At $t \approx \SI{8}{\second}$ and $t\approx \SI{11}{\second}$ an external force of magnitude $\SI{40}{\newton}$ acts for $\SI{1}{\second}$ on the robot right arm. 
The external force is estimated considering the Force Torque sensors mounted on the robot arms and the joint state~\cite{Nori2015}.
The MPC automatically compensates for the disturbance effect by adapting the location of the footstep with an average of $\SI{5}{\centi \meter}$ -- Figs.~\ref{fig:icub_walking}a and \ref{fig:icub_walking}b.
\section{Conclusions}
\label{sec:conclusions}
This paper contributes towards the development of an online Centroidal Momentum non-linear MPC for humanoid robots. 
The controller aims to generate feasible contact locations and wrenches for locomotion.
Differently from state-of-the-art architectures based on simplified models (e.g. LIPM), the proposed controller can be used to perform highly dynamic movements, such as jumping and running. Furthermore, the contact location adjustment is considered in the Centroidal dynamics stabilization problem, thus it is not required to design an ad-hoc block for this feature. We validate the controller with a simulation of one-leg and two-leg systems performing jumping and running tasks, respectively. 
The Centroidal MPC is also embedded in a three-layer position-based control architecture and tested on the humanoid robot iCub. The proposed strategy prevents the robot from falling while walking and pushed with external forces up to \SI{40}{\newton} for 1 second applied at the robot arm.
\par
As future work, we plan to extend the MPC to consider also the contact timing adjustment, thus increasing the robustness properties against unpredictable external disturbances. Furthermore, to better test the presented controller, we plan to extend the whole-body control layer to cope with non-complanar contact scenarios, i.e.~\cite{Caron2017ZMPConstraints}. 
In addition, we also plan to validate the architecture on a torque-based walking controller architecture~\cite{Romualdi2020ARobots}. Here the performances of the low-level torque control along with the base estimation capabilities become crucial to obtain successful results.
A possible future research direction is to embed the controller in an architecture that considers the interaction with a compliant environment~\cite{Romualdi2021ModelingControl}. In this context, it is pivotal to ensure the tracking of the desired contact forces. To achieve this objective, we may perform a dynamic extension of the system~\eqref{eq:centroidal_dynamics_discretized} considering the contact forces as a state of the dynamical system and controlling their derivative~\cite{Gazar2021JerkFeedback}.
Finally, to improve the overall time performance, we are planning to warm start the non-linear optimization problem with the result of a human-like trajectory planner~\cite{Viceconte2022ADHERENT:Robots}.

\balance

\bibliography{locomotion}
\bibliographystyle{IEEEtran}

\end{document}